\documentclass[preprint,3p,12pt]{elsarticle}
\usepackage{graphicx}
\usepackage{tikz}
\usepackage{amsmath}
\usepackage{amssymb}
\usepackage{xcolor}
\usepackage{mathtools}
\usepackage{url}
\usepackage{subcaption} 
\usepackage[utf8]{inputenc}
\usepackage{float}
\usepackage{booktabs}
\usepackage{tabularx}
\usepackage{natbib} 
\usepackage{algorithm}
\usepackage{algpseudocode}
\usepackage{setspace}
\newcommand{\mycomment}[1]{}
\usepackage[capitalise]{cleveref}
\begin{document}
\begin{frontmatter}
\renewcommand{\topfraction}{0.95}
\renewcommand{\bottomfraction}{0.95}
\renewcommand{\textfraction}{0.05}
\renewcommand{\floatpagefraction}{0.9}

\title{Parameter-Efficient Conditioning for Material Generalization in Graph-Based Simulators}
\author[label1]{Naveen Raj Manoharan}
\ead{naveenrajmanoharan@utexas.edu}

\author[label2]{Hassan Iqbal}
\ead{hassan.iqbal@utexas.edu}

\author[label1,label2]{Krishna Kumar}
\ead{krishnak@utexas.edu}

\affiliation[label1]{organization={Department of Civil, Architecture and Environmental Engineering}, addressline={The University of Texas at Austin},country={USA}}
\affiliation[label2]{organization={The Oden Institute for Computational Engineering and Sciences}, addressline={The University of Texas at Austin},country={USA}}

\begin{abstract}

Graph network–based simulators (GNS) have demonstrated strong potential for learning particle-based physics (such as fluids, deformable solids and granular flows) while generalizing to unseen geometries due to their inherent inductive biases. However, existing models are typically trained for a single material type and fail to generalize across distinct constitutive behaviors, limiting their applicability in real-world engineering settings. Using granular flows as a running example, we propose a parameter-efficient conditioning mechanism that makes the GNS model adaptive to material parameters. We identify that sensitivity to material properties is concentrated in the early message-passing (MP) layers, a finding we link to the local nature of constitutive models (e.g., Mohr-Coulomb) and their effects on information propagation. We empirically validate this by showing that fine-tuning only the first few ($1$-$5$) of $10$ MP layers of a pretrained model achieves comparable test performance as compared to fine-tuning the entire network. Building on this insight, we propose a parameter-efficient Feature-wise Linear Modulation (FiLM) conditioning mechanism designed to specifically target these early layers. This approach produces accurate long-term rollouts on unseen, interpolated or moderately extrapolated values (e.g., up to $2.5$\textdegree~for friction angle and $0.25$ kPa for cohesion) when trained exclusively on as few as $12$ short simulation trajectories from new materials, representing a $3$-fold data reduction compared to baseline multi-task learning method. Finally, we validate the model's utility by applying it to an inverse problem, successfully identifying unknown cohesion parameters from trajectory data. This approach enables the use of GNS in inverse design and closed-loop control tasks where material properties are treated as design variables.

\end{abstract}




\end{frontmatter}

\doublespacing
\section{Introduction}
\label{sec:introduction}

\begin{figure*}[ht!]
    \centering
    \includegraphics{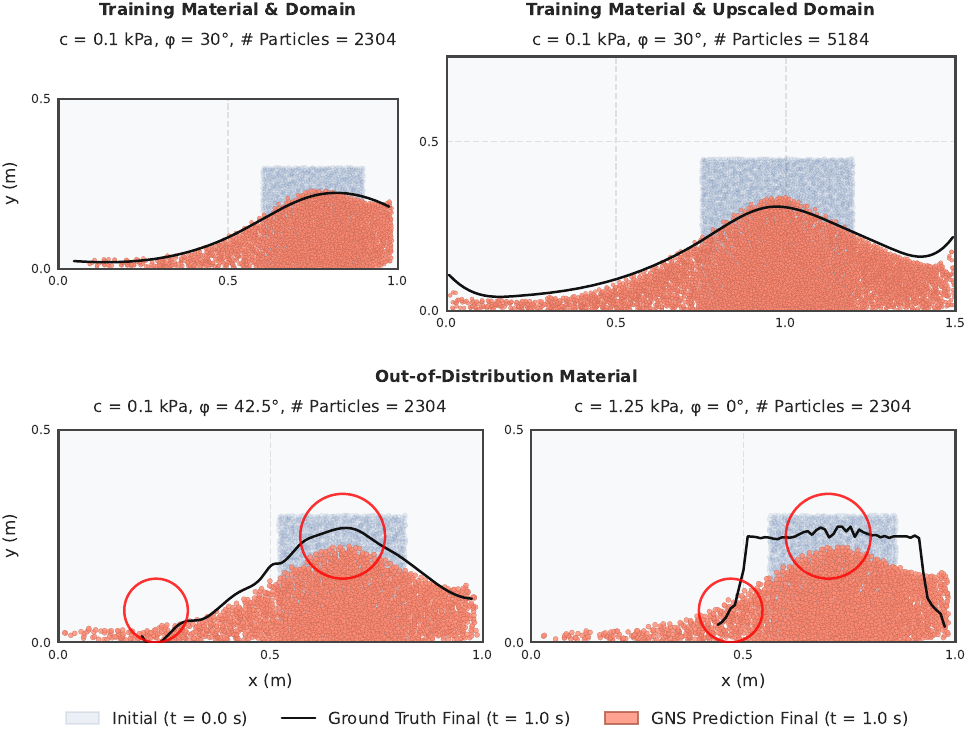}
    \caption{ GNS model trained on a granular material (friction angle, $\phi = 30^\circ$, $c = 0.1 kPa$) successfully generalizes to larger domain with a higher particle count, but fails to accurately predict dynamics for unseen material properties ($\phi = 42.5^\circ$ and $c = 1.25 kPa$). Ground truth trajectories are computed using a high-fidelity Material Point Method (MPM).}
    \label{fig:inductive_bias}
\end{figure*}

Particle-based physics simulators represent the state of a system (such as fluids, granular materials, and deformable or rigid bodies) as discrete particles, preserving physical properties through their interactions. Graph neural networks (GNNs) have shown strong potential for learning these particle dynamics effectively \cite{sanchez2018graph,sanchez2020learning,choi2024graph,hernandez2022thermodynamics}. Rather than learning in a purely black-box manner, graph network–based simulators (GNS) incorporate sparsity and physical invariances directly into their architectures, thereby simplifying the underlying learning problem \cite{battaglia2018relational,jegelka2022theory}. These inductive biases enable strong generalization to previously unseen geometries and scales.
The present work uses granular flow as a running example. As shown in the top row of Fig.~\ref{fig:inductive_bias}, a GNS trained on a smaller domain accurately predicts granular flow over a longer time horizon (400 time steps) in a larger domain without retraining. It is pertinent to note that this example involves a particle-based system with dynamic topology, where graph connectivity (i.e., particle interactions) is recomputed at every time step. This evolving connectivity makes the learning problem inherently more challenging than GNN applications with fixed topologies, such as mesh-based simulations \cite{pfaff2020learning}.  

In practice, these GNS models are typically trained and validated for single material type, thus give arbitrarily poor predictions in out-of-distribution (OOD) materials. For instance, the bottom row in Fig.~\ref{fig:inductive_bias} shows a learned GNS fails to accurately predict when internal friction angle or cohesion is different. A recent review \cite{zhao2024review} noted that most GNS works in mechanics omit material properties from their inputs, and have minimal material variations within their training datasets. These limitations stem from the optimization and learning difficulties inherent in capturing complex, path-dependent material constitutive responses, and the high computational cost of assembling datasets that span diverse material parameters. Such persistent challenges of GNS in maintaining reliability across material variations, more pronounced under limited labeled trajectory data setting, continue to hinder the broader adoption of learned models in computational mechanics \cite{yuan2022towards} and motivate the present work. 

To address the challenge, some approaches are to provide varying material properties as additional global or vertex features \cite{choi2024graph, choi2025differentiable}, or train a separate network for different materials. In both cases, offline data generation and training cost cannot be amortized in moderate to large particle systems, i.e. a large batch of high-fidelity labeled trajectory data are required corresponding to diverse material parameters and minimizing empirical loss involves a computationally intensive training process. Moreover, adding material properties as input features significantly increases optimization task complexity, and makes \textit{linear algorithmic alignment} more difficult with standard GNS' inductive biases. Hence, learning and extrapolation performance may suffer \cite{jegelka2022theory, xu2020neural}. Lastly, merely designing a deeper graph network (in hopes to be able to jointly learn varying material dynamics) has been shown to lead to over-smoothing \cite{rusch2023survey} and overfitting \cite{lecun2015deep} in other applications, especially in limited data regime.

Transferability and generalization of pretrained graph networks has been explored in applications such as molecular dynamics, drug discovery, quantum mechanics etc. Existing graph domain adaptation strategies often involve transfer learning by fine-tuning the final ``readout'' layers of a network \cite{buterez2024transfer} or by using the GNN's final latent embeddings as inputs for other models, such as kernel mean embeddings \cite{falk2023transfer}. Domain \textit{generalization} techniques include adversarially aligning the entire output embedding space \cite{zhang2019dane} or learning a single global latent vector to modulate the system's overall ODE dynamics based on environmental factors \cite{huang2023generalizing, forgione2023adaptation}. These approaches primarily treat domain shifts as a high-level problem, applying uniform adaptations to the model's final output representations or global state, but do not explicitly target how domain variations manifest in the local node and edge attributes that encode the underlying physics. We submit that jointly learning varying materials dynamics in GNS can be viewed as OOD generalization problem in node and edge attributes. 

Outside of graph neural networks, machine-learning approaches are employed to learn or adapt constitutive material models with measurements or numerical data \cite{wang2020learning,ma2023learning, lourencco2024indirect, moya2023thermodynamics}. Additionally, \textit{multi-task learning} has been used to distill a ``common core'' for recurrent neural network models of plasticity, enabling a single base model to capture a range of material behaviors \cite{heidenreich2024rnnplasticity}. However, such methods have limited generalization to unseen geometries in a zero-shot (retraining-free) manner. Lastly, existing works focused on \textit{meta-learning} parametric dynamical systems with neural-ODE representations (see \cite{kirchmeyer2022generalizing} and references therein), typically train a model from scratch using data from all ``source'' environments, each representing different system parameter (material properties in the case of present work). These approaches are data-hungry, and best suited for lower dimensional systems where data generation is inexpensive. 

In contrast, our approach considers a different, practical scenario for particle-based problems. We start with a learned GNS predictor \textit{pretrained} on a single source environment and formulate conditioning method that use a small set of auxiliary environments to achieve a learned model that is parametric to not just the state but also material parameter space. For pretraining, data can be generated for a single known material by experts via expensive high-fidelity material point methods (MPM) or though lab experiments (as is standard practice in GNS literature). The pretrained model has learned particle behaviors (momentum conservation) and the knowledge should be exploited in generalizing to varying materials. As a result, our domain generalization problem is reframed from a multi-domain joint training regime to one of targeted adaptation, where a limited dataset from a few material types is used to efficiently train a conditioning mechanism that adapts the pre-trained model to new constitutive behaviors, eventually allowing transferability in a zero-shot manner. 

Our approach is built upon parameter-efficient fine-tuning (PEFT), primarily employed in computer vision and natural language processing (NLP), with a key difference that existing PEFT for graph neural networks distribute adaptations throughout the entire model architecture while we design modulations on identified ``material specific'' features. For instance, tunable adapter modules are introduced into each layer to run in parallel with the frozen network components in \cite{li2024adaptergnn}, while in the specialized domain of equivariant models, fine-tuning is concentrated on the iterative ``interaction blocks'' that form the network’s core \cite{wangelora}. Similarly, other work has focused on efficient test-time adaptation that prevents catastrophic forgetting \cite{niu2022efficient}. The shared philosophy behind these approaches is that effective adaptation requires tunable modules distributed across the network's full depth. In parallel, conventional transfer learning wisdom in computer vision, and even some works in GNN \cite{zhang2019dane}, has suggested that generalization is primarily a high-level abstraction problem and only the deeper layers need to be adapted to the target tasks.  

Current work reveals a departure from this perspective in that we show that the most domain (material) sensitive computations in GNNs occur within the first few message passing (MP) layers. Our empirical analysis, presented later in the paper, traces this to effects of material variations that are predominantly captured within local particle interactions. We hypothesize this is consistent with continuum mechanics, as material-specific constitutive models (e.g., Mohr-Coulomb) are fundamentally local. These models define the acoustic wave speed ($c_p$), which in turn limits the speed of information propagation via the CFL condition. The GNS' early layers, with their inherently local receptive fields, are the natural architectural component for learning this local, material-dependent physics. Building on this insight, we design a feature-wise conditioning mechanism \cite{perez2018film, dumoulin2018feature-wise} that specifically targets these early, local MP layers. The conditioning mechanism reduces sample complexity and improves prediction accuracy across unseen material parameters.

Our contributions are as follows:
\begin{itemize}
    \item We identify that domain sensitivity in GNS is concentrated in the early MP layers, with links to local, material-dependent physics.
    
    \item We propose a parameter-efficient conditioning mechanism that targets these early layers, enabling generalization across a continuous spectrum of material properties.
    
    \item We generate a high-fidelity trajectory dataset of granular flows with varying cohesion and internal friction angles. The data will be publicly released upon publication.
    
    \item We introduce a principled mechanism to select informative trajectory segments, accelerating model adaptation for new materials and reducing data requirements.
    
    \item We perform extensive validation on long rollouts, demonstrating that our approach generalizes effectively to unseen, interpolated and extrapolated material properties.
    
    \item We demonstrate the model's utility by applying it to an inverse problem: successfully identifying material properties from trajectory data.
\end{itemize}

The paper is organized as follows. Section \ref{sec:background} provides a background on the baseline MP GNS architecture, defines domain generalization in context of current work, and the constitutive model used in ground-truth data generation. Section \ref{sec:methodology} introduces proposed conditioning framework and the mechanism for informative data selection. Section \ref{sec:experimental_setup} outlines the experimental setup, including datasets, evaluation metrics, and baselines used for validation. Results and discussion is presented in section \ref{sec:result}. Finally, section \ref{sec:conclusion} concludes the paper.

\section{Background}
\label{sec:background}
\subsection{Message passing GNS}
\label{subsec:gns_description}

\begin{figure*}
    \centering
    \includegraphics[width=\linewidth]{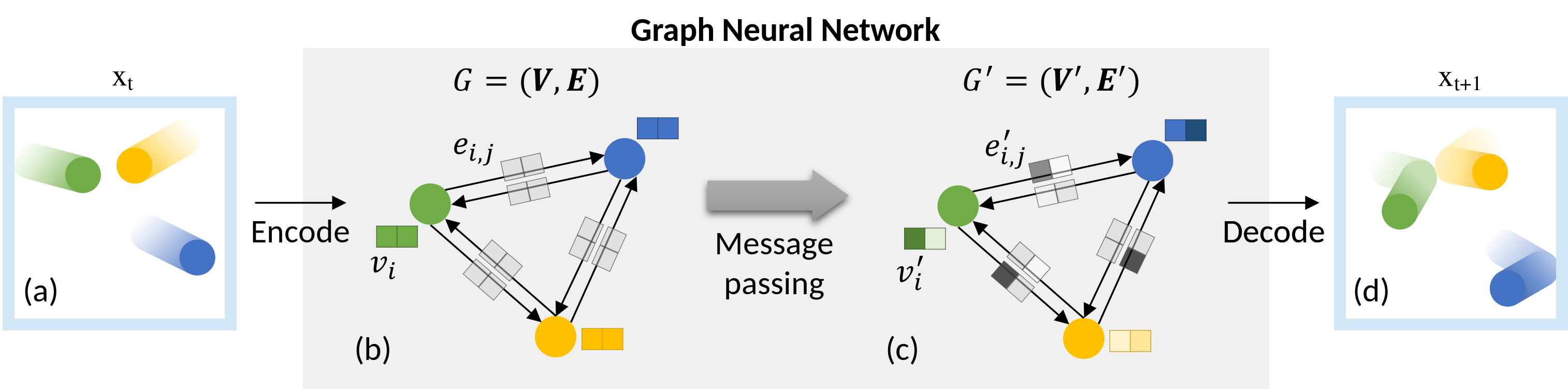}
    \caption{Schematic representation of a GNS model.}
    \label{fig:Gns-schematic}
\end{figure*}

Graph network-based simulators (GNS) are machine learning methods that leverage graph neural network architectures to model and simulate particle and fluid flows. They learn to represent the state of a physical system as a graph consisting of nodes and edges. Graph-based approaches offer permutation invariance, i.e. GNS outputs remain consistent regardless of the order in which nodes are presented, which is crucial for handling unordered data such as in particle systems. GNS learn local interactions between particles in the physical system via a data-driven loss on observed history. Once trained, GNS have demonstrated remarkable success in simulating complex physical dynamics over long time horizons with high accuracy, stability and computational efficiency \cite{pfaff2020learning, sanchez2020learning, choi2024graph}. 

A key advantage of GNS-based approaches is their strong generalization ability, which stems directly from their graph-based architecture that learns interactions within local neighborhoods. By modeling particle dynamics through pairwise relations, these systems capture both the shared physical interactions among particles within the inner domain and the collision dynamics at boundaries. This localized learning approach enables GNS to accurately predict behaviors in test scenarios that differ significantly from the training data, including environments with unseen obstacles, positions and velocities, or different spatial domain ranges.

The process of graph-based simulation is shown in \cref{fig:Gns-schematic} (modified from \cite{battaglia2018relational}). The current state of physical system $\boldsymbol{x_t}$ is represented as graph  $\Gamma = (\textbf{V},\textbf{E})$, information is propagated through the graph with MP and the updated graph $\Gamma'=(\textbf{V'},\textbf{E'})$ is decoded to give predicted state of the physical system. For example, in the case of granular soil flows, vertices and edges can represent grains and their directional interactions respectively, and the updated graph outputs the grains positions at the next time step.

Given node feature vectors $\boldsymbol{v}_i$ for all $i \in \mathcal{V}$ and edge feature vectors $\boldsymbol{e}_{i,j}$ for all $ {(i,j)} \in \mathcal{E}$, the graph $\Gamma^{(l)}$ at the $l$-th layer is updated through learned representation given as,

\begin{equation}
    \Gamma'^{(l)} = \underset{\boldsymbol{v}, \boldsymbol{e}}{\mathcal{GNN}}(\Gamma^{(l-1)})
\end{equation}

where $\boldsymbol{v}$ and $\boldsymbol{e}$ are node and edge embeddings respectively. We define the neighborhood indices of node $\boldsymbol{v}_i$ as $\mathcal{N}(i) = \{j \mid (v_j, v_i) \in \mathcal{E} \}$. Hence, the MP to update node $\boldsymbol{v}_i$ and edge $\boldsymbol{e}_{i,j}$ embeddings can be described as, 

\begin{subequations}
    \begin{align}
        &\boldsymbol{v}_i^{(l)} = \mathbf{f}_{\boldsymbol{\theta}}^{(l)}
        \left(
            \boldsymbol{v}_i^{(l-1)},
            \underset{j \in \mathcal{N}(i)}{\bigoplus} 
            \mathbf{g}_{\boldsymbol{\theta}}^{(l)}
            \left(
                \boldsymbol{v}_j^{(l-1)}, 
                \boldsymbol{v}_i^{(l-1)}, 
                \boldsymbol{e}_{i,j}^{(l-1)}
            \right)
        \right) \\
        &\boldsymbol{e}_{i,j}^{(l)}(w, v) = 
        \mathbf{g}_{\boldsymbol{\theta}}^{(l)}
        \left(
            \boldsymbol{v}_j^{(l-1)}, 
            \boldsymbol{v}_i^{(l-1)}, 
            \boldsymbol{e}_{i,j}^{(l-1)}
        \right).
    \end{align}
    \label{eq:GNS_equations}
\end{subequations}

where $\mathbf{f_\theta}$, $\bigoplus$ and $\mathbf{g_\theta}$ are embedding update function, the aggregate function and message generation function respectively. Here, $\mathbf{f_\theta}$ and $\mathbf{g_\theta}$ are learned functions represented by multi-layer perceptrons (MLPs) while the $\bigoplus$ can be any permutation invariant operator such as summation, mean or maximum \cite{zhao2025physical}. 

A given GNS model operates on a high dimensional latent space determined by the dimensions of node and edge embeddings. While these embeddings capture complex particle properties and interactions, the neural networks used to represent $\mathbf{f_\theta}$ and $\mathbf{g_\theta}$ are typically shallow rather than deep. This architectural choice is deliberate since shallow networks with sufficient width can effectively learn the necessary transformations for ``algorithmic alignment'' while maintaining computational efficiency. The key complexity in GNS comes not from network depth but from the iterative MP structure and the graph representation itself, which inherently models particle interactions. 

\subsection{Domain generalization}

\begin{figure*}[t]
    \centering
    \includegraphics[width=\textwidth]{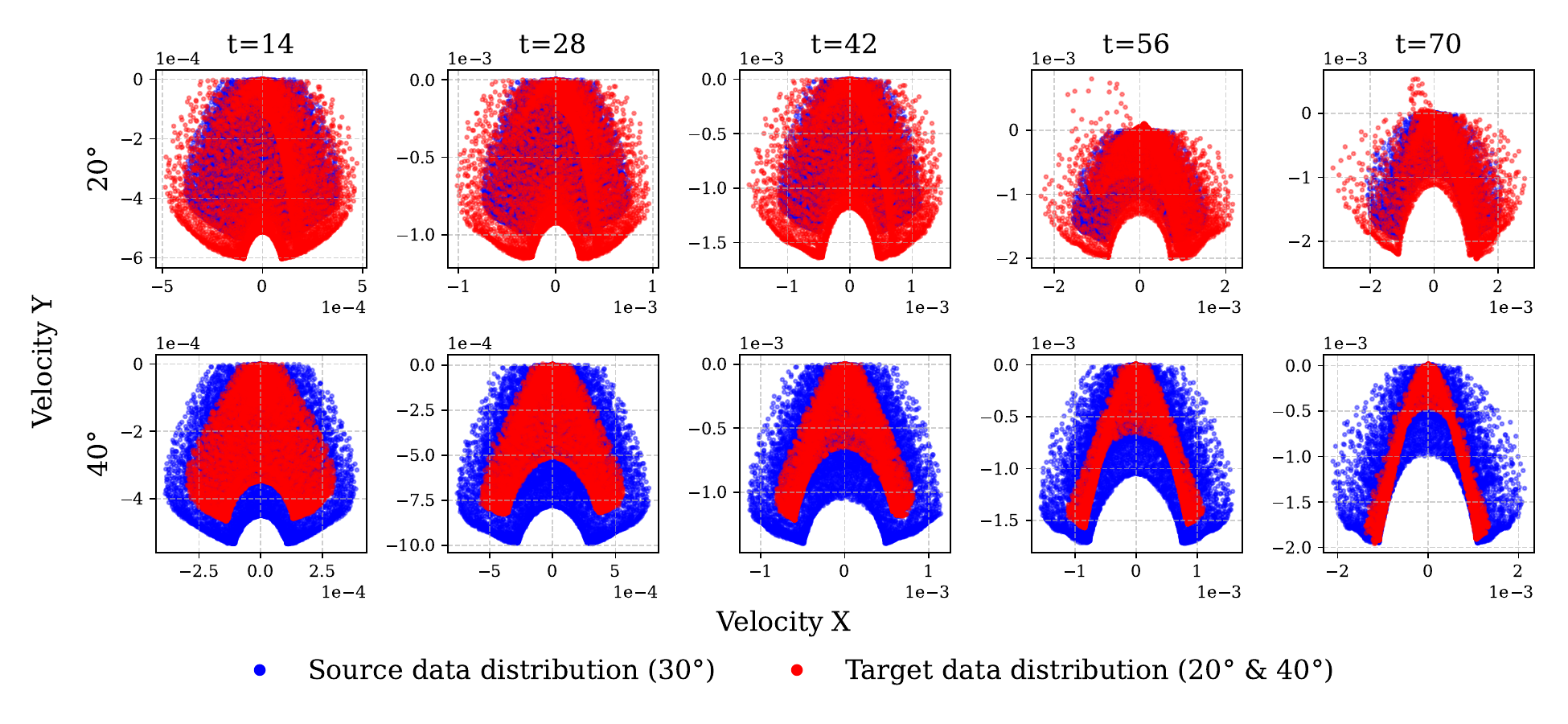}
    \caption{Out-of-distribution (OOD) evaluation of a GNS pretrained on granular collapses with an internal friction angle of $30^{\circ}$. The scatter plots show particle velocities in the $x$- and $y$-directions across different timeframes. Deviations between the pretrained model and unseen friction angles ($20^{\circ}$ and $40^{\circ}$) highlight the distribution shift}
    \label{fig:ood-proof}
\end{figure*} 

We formally define domain generalization in the context of this work. Let $\mathcal{E}$ denote the space of environments, where each environment $e \in \mathcal{E}$ is characterized by a joint distribution $P_{XY}^e$ over inputs $X$ and outputs $Y$. A learning algorithm has access to data from a collection of source environments $\mathcal{E}_{\text{train}} = \{e_1, \dots, e_M\}$ and aims to learn a predictor $\underset{\boldsymbol{v}, \boldsymbol{e}}{\mathcal{GNN}} : X \to Y$ that performs well on previously unseen environments $e_{\text{test}} \in \mathcal{E}_{\text{test}}$ with $\mathcal{E}_{\text{test}} \cap \mathcal{E}_{\text{train}} = \emptyset$. Performance is measured by the population risk
\[R_{e_{\text{test}}}(h) = \mathbb{E}_{(x,y)\sim P_{XY}^{e_{\text{test}}}} \big[\ell(\underset{\boldsymbol{v}, \boldsymbol{e}}{\mathcal{GNN}}(x),y)\big].\]
where $l:Y\times Y\rightarrow R$ is a specified loss function. The key challenge in domain generalization is to minimize this risk without access to samples from $\mathcal{E}_{\text{test}}$ during training. Fig.~\ref{fig:ood-proof} illustrates a data distribution shift between two environments, i.e. trajectory distributions differ substantially when sampled from different internal friction angles. A similar phenomenon can be observed for varying cohesion. The relationship is described by the governing Mohr-Coulomb constitutive model. Present work considers the running example where $e$ can be considered as parametric material cohesion or internal friction angle; $X$ and $Y$ respectively to be the particle states and their one step predictions, i.e. $\underset{\boldsymbol{v}, \boldsymbol{e}}{\mathcal{GNN}}:X_k\rightarrow X_{k+1}$.

\subsection{Mohr-Coulomb model}
\label{subsec:mohr}
The Mohr–Coulomb criterion postulates that yielding takes place when the shear stress at a point within the material attains a magnitude that varies linearly with the normal stress acting on the same plane. For a general stress state, the yield function is more conveniently expressed using the three stress invariants as follows, 

\begin{subequations}
\begin{align}
    F &= R_{mc}q + p' \tan \phi - c \\
    R_{mc}(\theta, \phi) &= \frac{1}{\sqrt{3}\cos\phi} \sin\!\left(\theta + \frac{\pi}{3}\right)
    + \frac{1}{3} \cos\!\left(\theta + \frac{\pi}{3}\right) \tan \phi
\end{align}
\end{subequations}

where $p^{\prime}$ is the effective mean pressure, $q$ is the magnitude of deviatoric stress, $\theta$ is the Lode's angle, $\phi$ is the effective friction angle, $c$ is the effective cohesion. We employ Mohr-Coulomb in the ground-truth trajectory data generation. 

\section{Methodology}
\label{sec:methodology}

\subsection{Layer sensitivity analysis}
\label{subsec:layer_sensitivity}

\begin{figure}[t]
    \centering
    \includegraphics[width=3.25in]{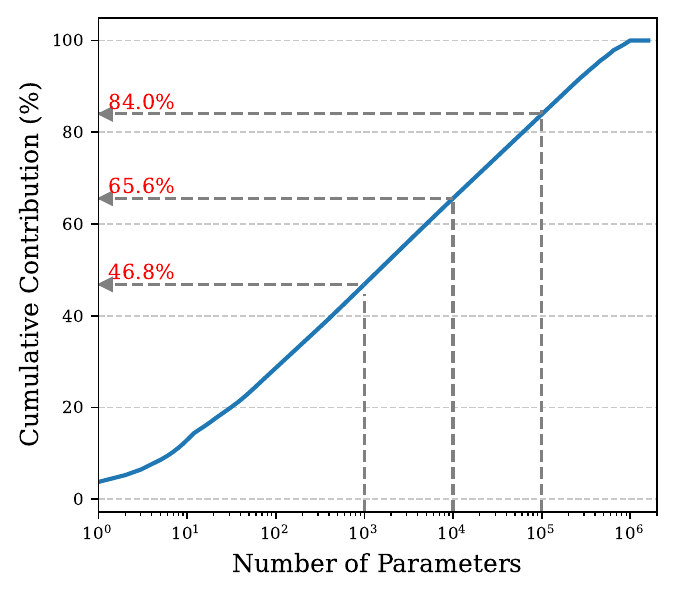}
    \caption{Cumulative distribution function of parameter changes by component when fine-tuning for the target friction angle. The steep rise for the processor indicates that few parameters dominate the total update magnitude.}
    \label{fig:sensitivity_cdf}
\end{figure}

The GNS models particle interactions through iterative MP updates. These layers are central to capturing material effects, yet they also comprise over 90\% of the model’s parameters in our implementation. Unlocking all layers for adaptation or generalization can be computationally inefficient and may lead to overfitting in low-data regimes. Therefore, we investigate the layer-wise sensitivity of the learned model to understand how the GNS model adapts to unseen material properties. In particular, our goal is to investigate whether a few layers of the architecture are more responsive to variations in material behavior, such as changes in internal friction ($\phi$) or cohesion ($c$).

\begin{figure*}[ht!]
    \centering
    \includegraphics{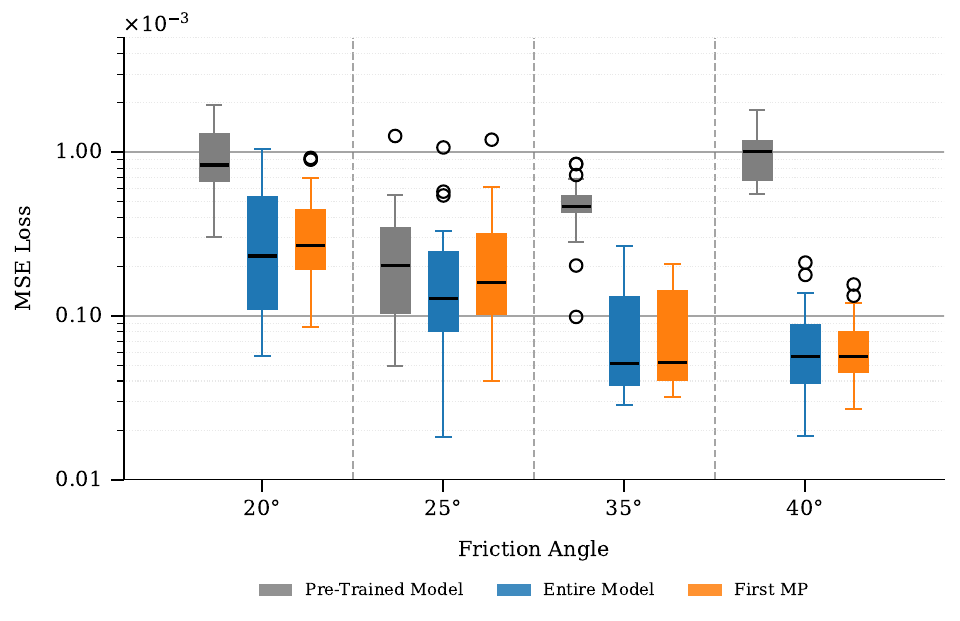}
    \caption{
        Test loss comparison for different fine-tuning configurations: full processor, and individual MP layers (first, fifth, and last). Despite comprising a small fraction of the total model parameters, fine-tuning a single MP layer achieves accuracy comparable to updating the full processor, indicating that material-specific adaptation is concentrated in specific subcomponents of the network.
    }
    \label{fig:mp_loss_comparison}
\end{figure*}

The GNS architecture provides a structural basis for this investigation. The MP process, defined in eq.~\eqref{eq:GNS_equations}, is iterative. At the initial layer ($l=1$), a node $\boldsymbol{v}_i$ aggregates messages $\mathbf{g}_{\boldsymbol{\theta}}^{(l)}$ only from its immediate, 1-hop neighbors $j \in \mathcal{N}(i)$. With each subsequent layer, the receptive field expands; information from a 2-hop neighbor must pass through a 1-hop neighbor at $l=1$ to reach $\boldsymbol{v}_i$ at $l=2$. Consequently, deeper layers aggregate information from progressively larger neighborhoods, capturing higher-level and more global kinematic structures. This architectural property aligns with the physics of granular materials. Material-dependent behaviors, such as the yielding described by the Mohr-Coulomb criterion (see section \ref{subsec:mohr}) used to generate the ground-truth data, are fundamentally local phenomena. This criterion is defined by local stress states, particle-to-particle contact forces, and material properties like the friction angle and cohesion.

A division of representational roles is therefore expected to emerge. Initial MP layers, which operate directly on local geometric features (e.g., pairwise distances, relative displacements), are positioned to predominantly capture these material-specific, local-contact responses. Conversely, the deeper layers are expected to capture higher-level, material-invariant global kinematics.

This expectation is confirmed by two empirical tests. First, we track parameter update magnitudes during fine-tuning to a new material with all layers unlocked. We found that material-specific adaptation is highly concentrated. As shown in \cref{fig:sensitivity_cdf}, a small subset of parameters (approximately 6.28\%) accounts for nearly 85\% of the total update magnitude. Second, we conduct ablation studies by fine-tuning the pretrained model to materials with different friction angles while updating only specific layers. As shown in \cref{fig:mp_loss_comparison}, fine-tuning just a single MP layer yields test losses that are nearly indistinguishable from fine-tuning the entire model. Additional results supporting this observation are provided in~\ref{app_ablation}. This analysis reinforces that the initial MP layers offer an efficient target for adaptation without disrupting the model’s general dynamical representations. In line with this, we confine all subsequent generalization experiments to these initial layers.

\subsection{Physics-guided trajectory sampling}
\label{subsec:data-selection}

\begin{figure*}[ht!]
    \centering
    \includegraphics[width=6.5in]{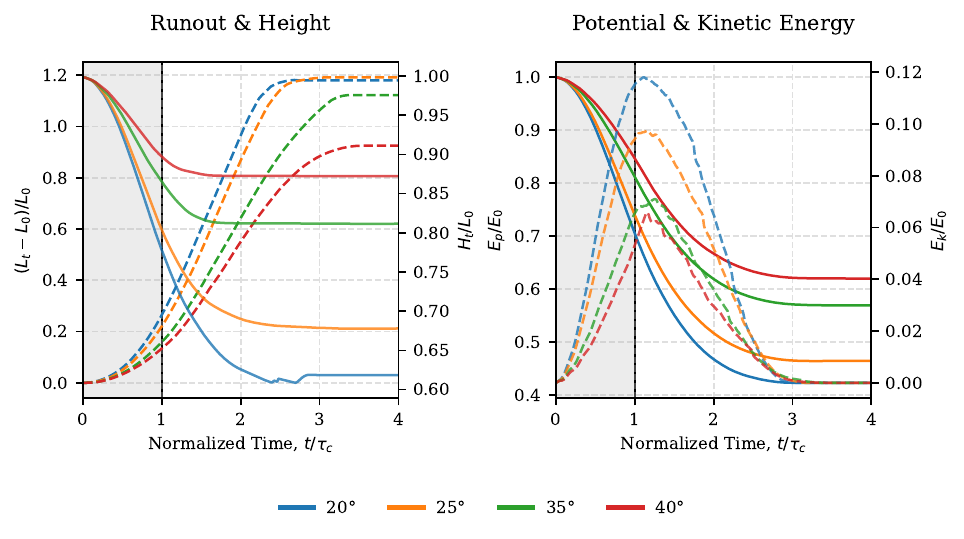}
    \caption{
        Flow dynamics of granular column collapse for varying internal friction angles.
        \textbf{(a)} Evolution of normalized runout distance and column height over time.
        \textbf{(b)} Normalized potential and kinetic energy during collapse. The early-time window ($0 \leq t/\tau_c \leq 1$), culminating in peak kinetic energy, captures the most information-rich, material-dependent behaviors crucial for model adaptation.
    }
    \label{fig:collapse-dynamics}
\end{figure*}

Generating high-fidelity trajectories for new materials is a significant computational bottleneck (see section~\ref{subsec:times}). We employ a physics-guided data selection strategy that drastically reduces this cost by identifying the most physically-informative portion of the trajectory. Our approach is guided by the underlying physics, as opposed to model-centric ``active learning'' heuristics (e.g., predictive uncertainty \cite{settles2009active}).

We consider a normalized time for a granular column collapse defined as $t/\tau_c$, where $ \tau_c = \sqrt{{H_0}/{g}}$, with $H_0$ being the initial column height and $g$ being the acceleration due to gravity. This characteristic timescale provides a reference for the flow evolution, with full mobilization of the granular material typically occurring around $t/\tau_c \sim 1$. Prior studies \cite{choi2024graph} indicate that the majority of the dynamic behavior with the strongest dependence on the internal friction angle occurs within this early-time window. By restricting fine-tuning to this early information-rich window, our method significantly reduces the dataset size and simulation cost while preserving the critical dynamics needed for effective adaptation. 

\begin{figure*}[t]
    \centering
    \includegraphics[width=6.5in]{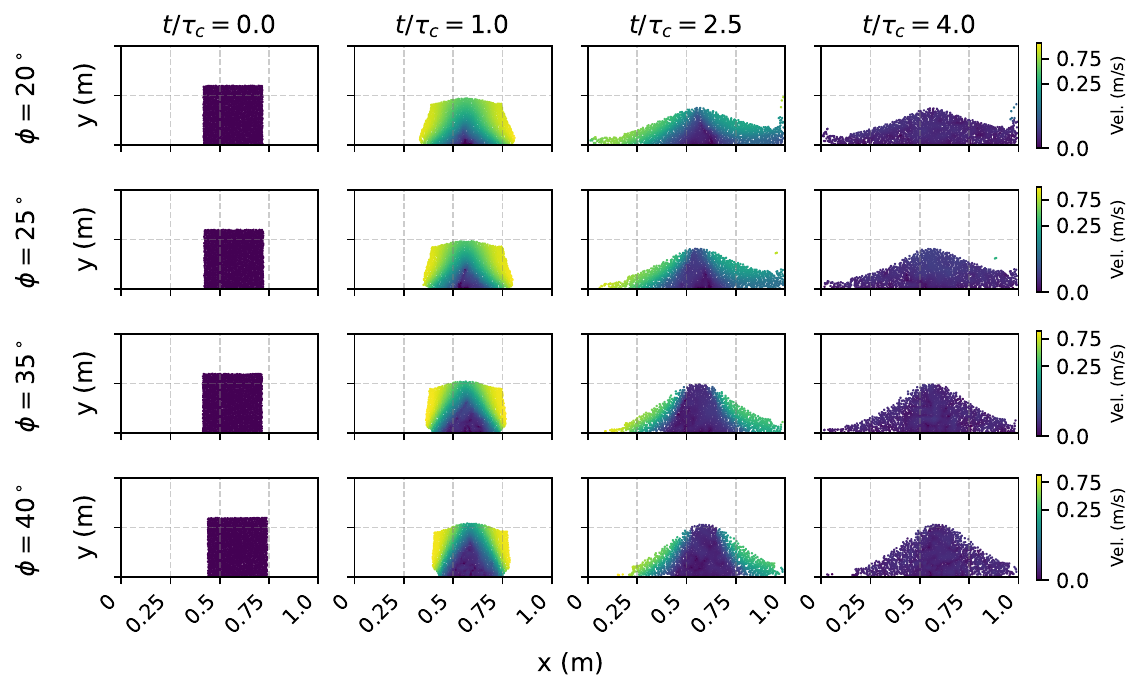}
    \caption{Qualitative comparison of granular column collapse dynamics across different internal friction angles ($\phi$) and normalized times ($t/\tau_c$). Each row corresponds to a distinct internal friction angle, and each column to a time snapshot. Particle positions are colored by velocity magnitude. The visualization highlights that the most distinct, material-dependent flow patterns emerge in the early mobilization phase ($t/\tau_c \approx 1.0$).}
    \label{fig:rollout_comparison}
\end{figure*}

To empirically substantiate our claim, we analyze the collapse dynamics across varying internal friction angles, as shown in \cref{fig:collapse-dynamics}. The evolution of normalized runout and height (\cref{fig:collapse-dynamics}a) reveals the primary kinematics, while the energy profiles (\cref{fig:collapse-dynamics}b) offer deeper insight into the underlying dynamics. During the initial phase, potential energy is rapidly converted into kinetic energy as the material descends and disperses. The kinetic energy peaks near $t/\tau_c \approx 1$, signaling that the granular material has reached full mobilization. Beyond this point, the system enters a dissipative regime dominated by basal friction, where the flow gradually decelerates into a quasi-static state. It is within the early window, $0 \leq t/\tau_c \leq 1$, that the material-dependent behaviors are most pronounced.

We therefore postulate that the early mobilization phase of the granular collapse ($t/\tau_c < 1$, representing roughlt 30\% of the total trajectory) encapsulates sufficient information for the model to infer the complete dynamics of the system. In the qualitative comparison shown in \cref{fig:rollout_comparison}, this interval corresponds to the first two columns of snapshots. The GNS is fine-tuned exclusively using this truncated portion of each trajectory, yet it achieves testing accuracy comparable to that obtained with full-length trajectories across all friction angles. The model’s ability to extrapolate from this limited window and accurately predict the system’s subsequent evolution up to the final quasi-static state ($t/\tau_c = 4.0$), shown in the last column, demonstrates that the transient mobilization phase encodes the dominant constitutive cues governing the later flow and deposition. This result underscores that data-efficient adaptation can be realized without sacrificing predictive fidelity, provided the selected samples capture the physically informative regime of the dynamics.

\subsection{Domain generalization with FiLM}
\label{subsec:film}

To generalize a pre-trained GNS across varying material parameters, such as internal friction angles or cohesion in granular media, we adopt Feature-wise Linear Modulation (FiLM)~\cite{dumoulin2018feature-wise} as a lightweight and effective conditioning mechanism. Given an intermediate feature vector ${h} \in \mathbb{R}^d$ at a layer within a neural network, FiLM modifies it using learned scale and shift parameters:
\begin{equation}
    \text{FiLM}({h} \mid {\gamma}, {\beta}) = {\gamma} \odot {h} + {\beta}
\end{equation}
where ${\gamma}, {\beta} \in \mathbb{R}^d$ are computed from a small conditioning MLP, and $\odot$ denotes element-wise multiplication. The pretraining-conditioning paradigm avoids the computationally and data-intensive process of either training separate models for each condition or training a single multitask model from scratch. FiLM functions by dynamically adapting the pre-trained model's internal activations based on external, material-specific parameters. This approach effectively separates concerns: the core GNS retains its strong inductive biases for geometry generalization and momentum conservation, while the FiLM hypernetwork (a small MLP) learns to model the non-linear interrelationships of distinct constitutive models.

\begin{figure*}[t]
    \centering
    \includegraphics[trim=30 150 30 150, width=\textwidth]
    {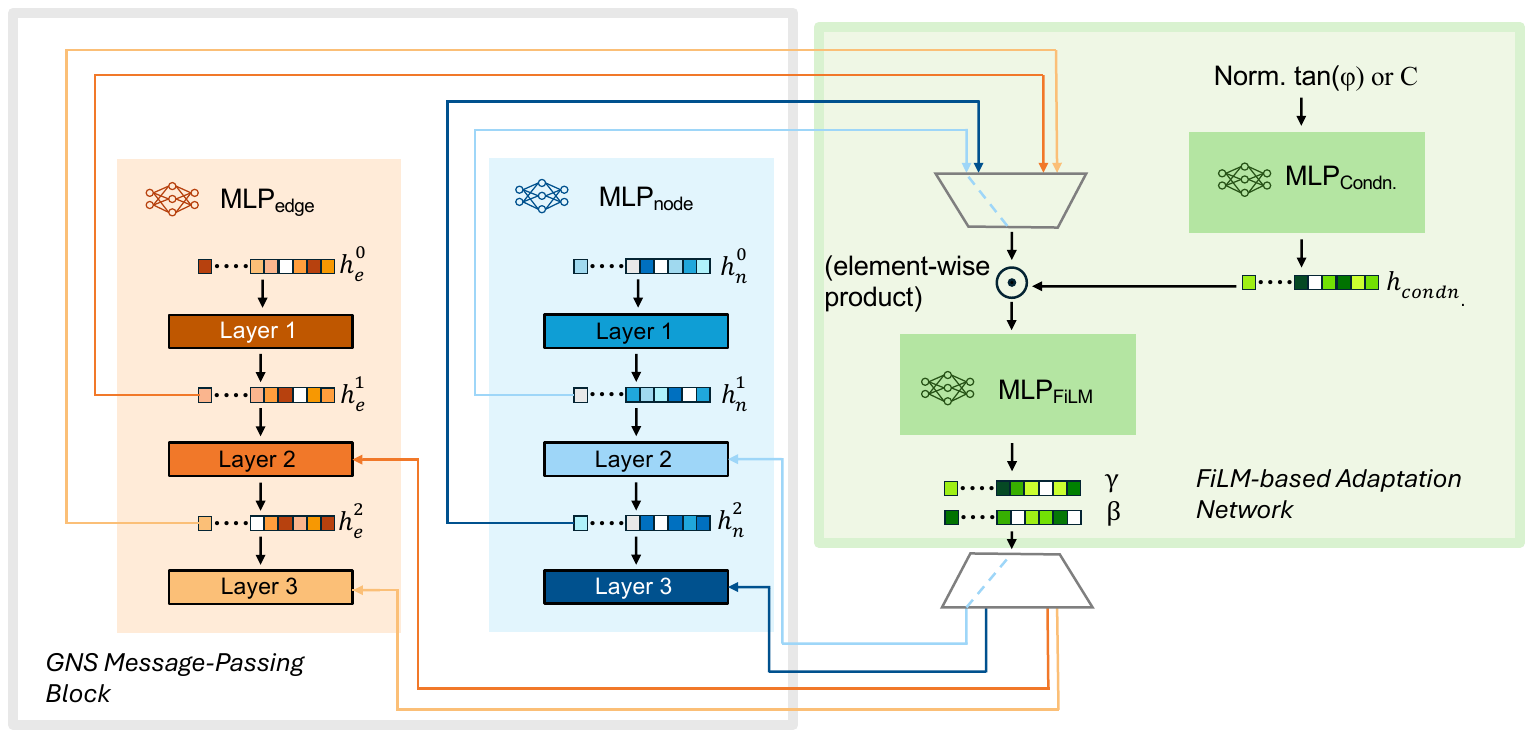}
    \caption{Schematic of FiLM-based adaptation in GNS. FiLM layers are applied to layers 2 and 3 of both the edge and node MLPs within the first MP block. The modulation parameters ${\gamma}^{(l)}$ and ${\beta}^{(l)}$ are generated by conditioning on the normalized material parameter (e.g., internal friction angle $\tan(\phi)$ or cohesion $c$) and the input activations to each layer. This enables material-aware and state-sensitive adaptation of the pre-trained GNS.}
    \label{fig:film_architecture}
\end{figure*}

A key design principle in our FiLM integration is that the modulation must be conditioned on both the material properties and the evolving particle state. Conditioning purely on a static parameter (like the friction angle $\phi$ or cohesion $c$) is insufficient, as the material's response is highly path-dependent. This can also be intuitively understood through the constitutive update step in the CB-Geo MPM solver~\cite{kumar2019scalable} used for ground truth data generation in present work. Here, stress updates depend on both the material's constants (e.g., $\phi, c$) and its evolving stress invariants (like $p'$, $q$, and $\theta$ from section~\ref{subsec:mohr}). The learned simulator must also capture this path-dependency. Therefore, by designing the FiLM conditioning generator on both the material parameter and the intermediate activations $h^{(l-1)}$ (which encode the instantaneous mechanical state in latent space) during MP iterations at each time step, the modulation parameters $(\gamma^{(l)}, \beta^{(l)})$ can adapt the latent dynamics of the pre-trained GNS in a manner consistent with classical constitutive updates. 

Our architecture (illustrated in \cref{fig:film_architecture}) builds on a pre-trained GNS model described in Section~\ref{subsec:gns_description}. We apply FiLM into the first few MP blocks of the processor. Each block consists of two 3-layer edge and node MLPs that are shared among all particles ($\mathbf{f_\theta}$ and $\mathbf{g_\theta}$ in eq.~\eqref{eq:GNS_equations}). The conditioning specifically targets the layers $2$ and $3$ of both the node and edge MLPs. This choice is guided by architectural constraints: as outlined in Section~\ref{subsec:gns_description}, the first layers of these MLPs receive differently sized inputs, making it infeasible to apply FiLM uniformly. In contrast, layers 2 and 3 share a consistent latent dimensionality and lie at a depth where modulation remains both expressive and stable. 

For each FiLM-modulated layer $l$, we begin by passing the relevant normalized material parameter, $\kappa$, through a small conditioning MLP. This parameter $\kappa$ represents the single value being generalized in a given experiment, for instance $\kappa = \text{norm}(\tan(\phi))$ for friction angle generalization or $\kappa = \text{norm}(c)$ for cohesion generalization. This output is then combined with the input activation $h^{(l-1)}$ via an element-wise product operation, which functions as a gating mechanism: the material-conditioned output selectively scales (emphasizes, suppresses, or inverts) the state-dependent activations $h^{(l-1)}$ to create a new, contextualized feature vector $z^{(l)}$ for subsequent processing. The resulting vector ${z}^{(l)}$ is passed to a separate FiLM MLP that outputs the affine transformation parameters, which are then used to modulate the pre-activation features of layer $l$:

\begin{subequations}
\label{eq:film_impl}
\begin{align}
    {z}^{(l)} &= \text{MLP}_\text{cond}(\kappa) \odot {h}^{(l-1)} \\
    {\gamma}^{(l)}, {\beta}^{(l)} &= \text{MLP}_\text{FiLM}\left( {z}^{(l)} \right) \\
    {h}^{(l)} &= \sigma\left( {\gamma}^{(l)} \odot ({W}^{(l)} {h}^{(l-1)} + {b}^{(l)}) + {\beta}^{(l)} \right)
\end{align}
\end{subequations}
where $\sigma$ denotes the non-linear activation function (e.g., ReLU), and ${W}^{(l)}$ and ${b}^{(l)}$ are the learnable weight matrix and bias vector of layer $l$, respectively.

\section{Experimental setup}
\label{sec:experimental_setup}
\subsection{Training data}

The pretrained model uses the dataset introduced by \cite{choi2024graph}, where the trajectories were generated using CB-Geo MPM code~\cite{kumar2019scalable}. The corresponding dataset is publicly available through DesignSafe-CI~\cite{kumar2023designsafe}. Each simulation models a granular column collapse under gravity using the Mohr–Coulomb constitutive law. Ground truth data is generated with fully numerical MPM solver with particle dynamics resolved over $1,000,000$ time steps with a time increment $\delta t = 10^{-6}$~s. To pretrain GNS, we uniformly subsample the trajectories every $2500$ time steps, resulting in $400$ time steps per trajectory with $\Delta t = 0.0025$~s. The pretraining dataset contains a total of $26$ such trajectories, each initialized with varying column heights, widths, and positions to capture a broad spectrum of granular flow dynamics. The key simulation parameters are summarized in Table~\ref{tab:mpm_parameters}.

\begin{table}[h]
\centering

\begin{tabular}{p{0.45\linewidth} p{0.45\linewidth}}
\toprule
\textbf{Property} & \textbf{Granular column collapse setup} \\
\midrule
Simulation boundary & $1.0 \times 1.0~\mathrm{m}$ \\
MPM element length & $0.025 \times 0.025~\mathrm{m}$ \\
Material point configuration & $25{,}600~\text{points}/\mathrm{m}^2$ \\
Granular mass geometry & $0.2 \times 0.2~\mathrm{m}$ and $0.3 \times 0.3~\mathrm{m}$ \\
Maximum number of particles & 2.3K \\
Simulation duration & 400 time steps ($\Delta t = 0.0025~\mathrm{s}$) \\
\midrule
\textbf{Material property} & \textbf{Mohr-Coulomb model} \\
Density & $1{,}800~\mathrm{kg/m^3}$ \\
Young’s modulus & $2~\mathrm{GPa}$ \\
Poisson’s ratio & $0.3$ \\
Friction angle & $30^{\circ}$ \\
Cohesion & $0.1~\mathrm{kPa}$ \\
Tension cutoff & $0.05~\mathrm{kPa}$ \\
\bottomrule
\end{tabular}
\caption{Simulation setup for pretraining~\cite{kumar2023designsafe}.}
\label{tab:mpm_parameters}
\end{table}

For the fine-tuning and multi-task learning experiments, all MPM parameters are kept identical to the pretrained setup except material properties. We use $12$ trajectories per material. The experiments are divided into two categories based on the material property being varied. The internal friction angle datasets isolate the effect of frictional resistance while keeping cohesion constant at $0.1~\mathrm{kPa}$. For the cohesion studies, the internal friction angle is fixed at zero to simulate purely cohesive behavior. All fine-tuning and multi-task learning experiments in this study use trajectories with an aspect ratio of 1.0, corresponding to the $0.3 \times 0.3~\mathrm{m}$ granular column configuration. 

\subsection{Baseline model}
\label{baseline}
To provide a reference for evaluating the effectiveness of fine-tuning and FiLM-based conditioning, we construct a baseline model trained using a conventional multi-material training strategy. The baseline architecture is identical to the pretrained GNS model, with the exception that an additional input dimension is included to represent the material property (e.g., friction angle). Similar to pretrained model, this model is first trained from scratch using trajectories corresponding to a single material configuration with an internal friction angle of $30^\circ$. After the model reaches convergence on this configuration, the training dataset is expanded to include trajectories from all other friction angles considered in this study. Training is then continued until convergence using the combined dataset. This sequential training procedure establishes a baseline that learns multiple material responses through standard training without the use of fine-tuning or feature-wise conditioning.

\subsection{Evaluation metrics}

We assess model performance using four complementary metrics: predictive loss, latent modulation smoothness, energy consistency, and computational efficiency. 

\paragraph{Test loss}
The predictive accuracy is measured using the mean squared error (MSE) between predicted and ground-truth accelerations. Each model variant (pretrained, fine-tuned, and FiLM-conditioned) is tested on 15 independent rollout trajectories. We visualize the resulting losses as box plots that display the median along with the 25\textsuperscript{th} and 75\textsuperscript{th} percentiles, which capture both the central tendency and variability across trajectories.

\paragraph{Normalized energy loss}
To verify physical consistency, we track the evolution of kinetic and potential energies computed from particle positions and velocities. For each rollout, we calculate the total kinetic energy as
\[
E_k(t) = \tfrac{1}{2} m \sum_i \|v_i(t)\|^2,
\]
and the total potential energy as
\[
E_p(t) = m g \sum_i h_i(t),
\]
where $m$ denotes the particle mass, $g$ is the gravitational acceleration, and $h_i$ is the particle height along the gravity axis. At time, $t = 0$, the system contains only potential energy, so we define the initial total energy as $E_0 = E_p(0)$. We then compute the normalized energy loss at time $t$ as
\[
\Delta E_{\mathrm{norm}}(t) =
\frac{|E_\mathrm{pred}(t) - E_\mathrm{true}(t)|}{E_0}.
\]
This dimensionless metric quantifies deviations from energy conservation, and lower values indicate greater physical consistency in the predicted rollouts.

\section{Result and discussion}
\label{sec:result}

\subsection{Generalization across friction angles}

\begin{figure*}[ht!]
    \centering
    \includegraphics[width=6.5in]{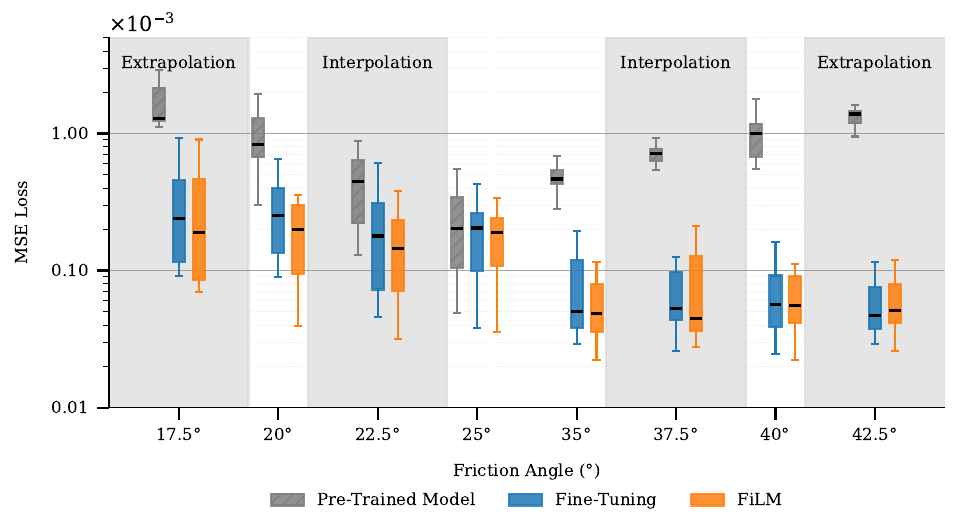}
    \caption{Test performance of FiLM-based multi-task GNS across different friction angles (17.5\textdegree–42.5\textdegree). FiLM consistently reduces error and variance compared to the pretrained baseline, demonstrating effective adaptation to different friction angles. Each model is evaluated over 15 test trajectories with varying initial conditions, and the results represent the average rollout loss over 400 time steps}
    \label{fig:friction_multitask}
\end{figure*}

We first evaluate the generalization of the conditioning framework to varying internal friction angles, applying FiLM adaptation only to the first MP block. The results are shown in \cref{fig:friction_multitask}. The models are trained on source angles of $20^\circ$, $25^\circ$, $35^\circ$, and $40^\circ$ and then tested on unseen trajectories corresponding to source, interpolated ($22.5^\circ$, $37.5^\circ$) and extrapolated ($17.5^\circ$, $42.5^\circ$) values. As expected, the pretrained baseline fails to adapt to these unseen angles. The fine-tuned models, while able to match a single new angle, is inherently not general, as this approach requires training and storing a separate model for each new condition. In contrast, the single FiLM-conditioned GNS generalizes smoothly across all unseen regimes. It maintains consistent accuracy and stable variance even under extrapolation, achieving an error comparable to specialized, fine-tuned models. The predictive errors of both the fine-tuned and FiLM-conditioned models are of the same order of magnitude as those obtained by the baseline model trained jointly on all friction angles (see~\ref{app_baseline}). This verifies the single adaptive framework's capacity to capture continuous material behavior through task-dependent modulation.

\begin{figure*}[ht!]
    \centering
    \includegraphics[width=6.5in]{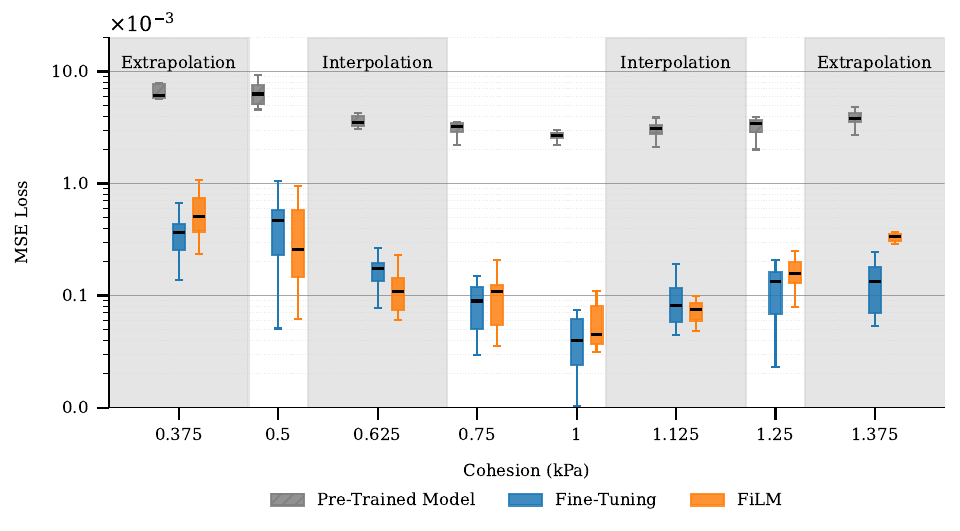}
    \caption{Test performance of FiLM-based multi-task GNS across different cohesion values (0.375–1.375~kPa). FiLM consistently reduces both error and variance compared to the pretrained baseline, demonstrating effective adaptation across cohesion regimes. Each model is evaluated over 15 test trajectories with varying initial conditions, and the results represent the average rollout loss over 400 time steps}
    \label{fig:cohesion_multitask}
\end{figure*}

\begin{figure*}[ht!]
    \centering
    \includegraphics{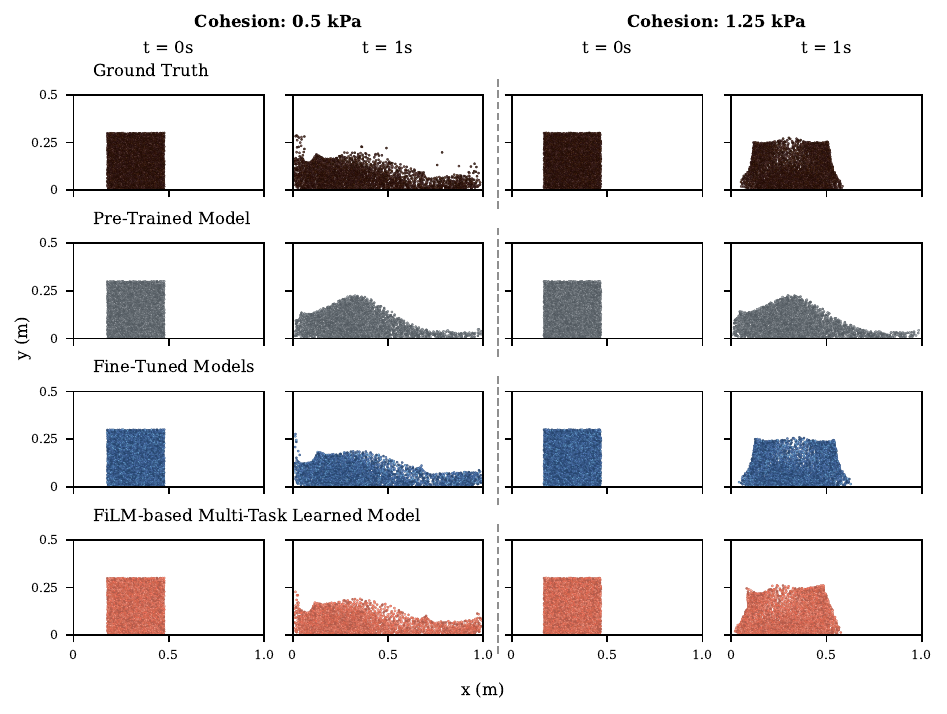}
    \caption{Qualitative comparison of granular collapse simulations under different cohesion levels (0.5~kPa and 1.25~kPa). Results are shown at the initial state ($t=0$~s) and final $t=1$~s for Ground Truth, Pre-Trained, Fine-Tuned, and FiLM models.}
    \label{fig:qualitative_comparison}
\end{figure*}

\subsection{Generalization across cohesion values}

To further challenge the framework, we test FiLM conditioning on cohesion, a regime that is physically distinct and empirically more complex than friction. The pretrained model was trained on a purely frictional ($\phi=30^\circ, c=0$) material, meaning it learned a purely stress-dependent strength model ($\tau \propto \sigma_n \tan(\phi)$). The cohesion task, which fixes $\phi=0$, introduces a stress-independent strength component ($\tau=c$), a physical mechanism the model has not seen before. This fundamental shift from a multiplicative to an additive term is reflected in our empirical finding: while friction required modulating only the first MP layer, cohesion required a more significant intervention, applying FiLM across the first five MP blocks. Nevertheless, the same external modulation network is used to generate the FiLM parameters for all layers.

Training is conducted over source cohesion values of $0.5$, $0.75$, $1.0$, and $1.25$~kPa. Model performance is then evaluated on unseen trajectories from source values as well as unseen interpolated ($0.625$, $1.125$~kPa) and extrapolated ($0.375$, $1.375$~kPa) regimes (shown in \cref{fig:cohesion_multitask}). The results mirror the friction experiments. The pretrained baseline fails to adapt to the unseen cohesion values. The fine-tuning baseline, while able to achieve low error for a single target value, is inherently not general, as it requires training and storing a separate model for each new cohesion parameter. In contrast, the single FiLM-conditioned GNS generalizes across all tested regimes. It maintains strong interpolation performance comparable to fine-tuned models trained specifically on tested values. Despite the expected degradation under extrapolation, continues to outperform the pretrained baseline by nearly an order of magnitude.

\cref{fig:qualitative_comparison} provides a qualitative validation. The pretrained model fails to distinguish between different material behaviors, producing nearly identical flow patterns across all cohesion levels. In contrast, both the fine-tuned models and the FiLM-conditioned model exhibit visibly improved agreement with the ground-truth simulations. The FiLM-conditioned model successfully captures the distinct physical behaviors (e.g., increased ``stickiness'') across all material strengths.

\subsection{PCA Analysis of FiLM parameters}
\label{sec:pca_film_parameters}

\begin{figure*}[ht!]
    \centering
    \includegraphics{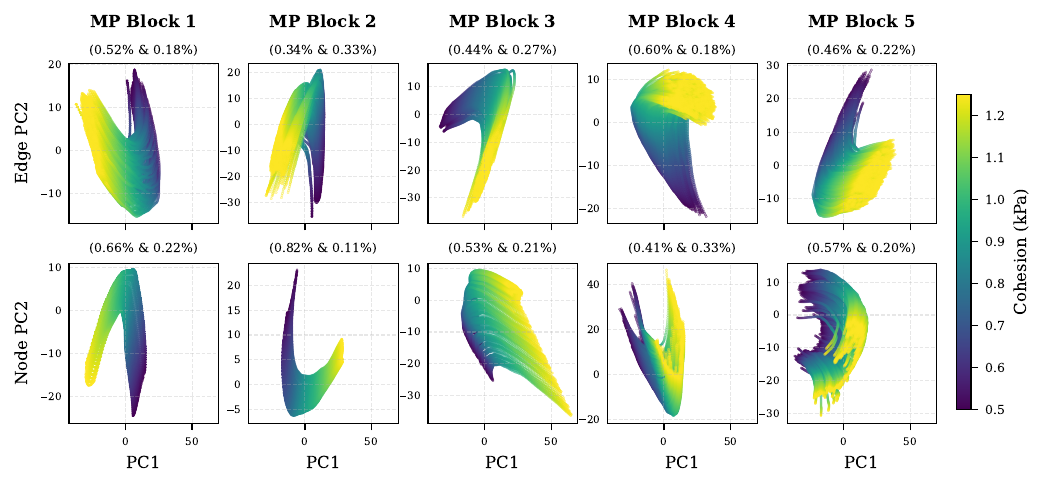}
    \caption{\textbf{Principal Component Analysis (PCA) of FiLM modulation parameters across MP blocks.}
    Each subplot shows the projection of FiLM parameters ($\gamma$, $\beta$) onto the first two principal components for edge and node functions across all MP blocks, color-coded by material property (e.g., cohesion in kPa). The values inside parentheses denote the percentage of variance captured by the first two principal components in each case. }
    \label{fig:pca_film_parameters}
\end{figure*}

 In Section~\ref{subsec:film}, we discussed that our goal in introducing FiLM-based conditioning was to generate modulation parameters that are sensitive both to the input material property (e.g., cohesion or friction angle) values and current latent state of the system. Current objective is to empirically analyze whether the learned affine transformation parameters ($\gamma$, $\beta$) indeed encode such material and state-dependent behaviors. To this end, for each FiLM-modulated MP block, we extract the corresponding $\gamma$ and $\beta$ values across a continuous range of material property values. These parameters are concatenated and then projected into a two-dimensional subspace defined by the first two principal components (PC1 and PC2) using Principal Component Analysis (PCA). In \cref{fig:pca_film_parameters}, we visualize the distribution of modulation parameters for both edge and node functions across all MP blocks, color-coded by the material property value (e.g., cohesion in kPa).

\begin{figure*}[ht!]
    \centering
    \includegraphics{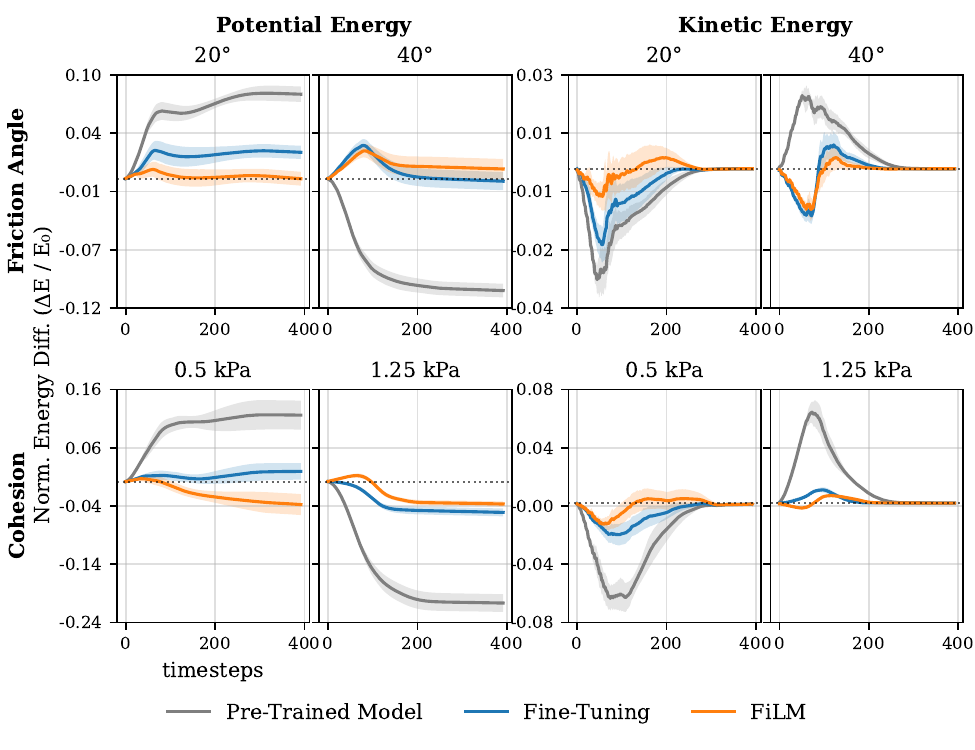}
    \caption{Normalized energy difference between GNS predictions and ground truth across rollout time steps for different friction angles. (a) shows the potential energy difference and (b) shows the kinetic energy difference. FiLM-based models maintain closer adherence to energy conservation.}
    \label{fig:energy_diff_plot}
\end{figure*}

Even though the FiLM conditioning network is only trained on $5$ source material parameter values, the FiLM modulations form continuous and smooth pattern in the reduced space, rather than discrete clusters, as a function of material values. As a notion of distance is present in the material properties and network activations inputted to conditioning network, continuity in its prediction in reduced space indicate that the conditioning network has learned a nonlinear smooth mapping between them. Secondly, this suggests that the conditioning network modulates the conditioned GNS' latent space in a physically meaningful manner that reflects the continuous variation of material behavior. This behavior contrasts with the findings of ~\cite{dumoulin2018feature-wise}, who reported distinct, task-specific clusters in the $\gamma$--$\beta$ space for visual reasoning models. Overall, this analysis demonstrates that conditioning strategy enables both compact parameterization and physically grounded generalization to unseen materials.

\subsection{Energy comparison}

Total energy is conserved through the interchange of potential, kinetic, and dissipation energies in column collapse scenario. To examine the physical consistency of the learned dynamics, we analyze the evolution of normalized potential and kinetic energy differences between the GNS predictions and the ground truth rollouts. \cref{fig:energy_diff_plot} shows the normalized energy differences over time for different friction angles. Each subplot illustrates the deviation in either potential or kinetic energy, computed at each rollout time step and normalized by the total initial energy. These plots help determine whether the model maintains physically plausible dynamics across different conditions. Notably, the FiLM-based model exhibits smaller energy deviations over time, supporting the claim that improvements in data-driven error metrics correspond to better adherence to the underlying physics.

\subsection{Computational time}
\label{subsec:times}
\begin{table}[ht!]
\centering
\caption{Computational time comparison for different training strategies.}
\label{tab:training_time}
\begin{tabular}{
    >{\raggedright\arraybackslash}m{0.18\columnwidth}
    >{\centering\arraybackslash}m{0.18\columnwidth}
    >{\centering\arraybackslash}m{0.16\columnwidth}
    >{\centering\arraybackslash}m{0.2\columnwidth}
}
\toprule
\textbf{Strategy} &
\textbf{Pretrain} &
\textbf{Fine-Tune} &
\shortstack{\textbf{Multi-Task}\\\textbf{(FiLM)}} \\
\midrule
Time & 192.96 h & 10 min & 2.65 h \\
Epochs     & 780      & 25     & 25     \\
\bottomrule
\end{tabular}
\end{table}
To evaluate the computational efficiency of different training strategies, we compare the total training time and number of epochs required for the pretrained model, fine-tuning, and FiLM-based multi-task learning. All experiments were conducted using a single NVIDIA A100 GPU with 40 GB of memory. Table~\ref{tab:training_time} summarizes the results.

The pretrained model was trained for a total of 780 epochs; however, we found that reasonable accuracy can typically be achieved after about 400 epochs, making the reported training time an upper bound. In comparison, both fine-tuning and multi-task learning reach convergence in practice at around 15 epochs, although we report results for 25 epochs as a consistent benchmark. Importantly, fine-tuning adapts the pretrained model to a single material with minimal computational cost, while the FiLM-based conditioning enables simultaneous generalization across a range of materials at only a modest increase in training time. 

An equally important aspect of computational efficiency lies in the cost of generating training trajectories. Each trajectory computation using CB-Geo MPM on 56 cores of Intel's Cascade Lake CPU with $128$ GB of memory on TACC Frontera requires approximately $104.05$ minutes. The pretrained model for a single material was trained on $26$ trajectories, while fine-tuning needs only $3$ trajectories, and FiLM-based multi-task learning requires 12 trajectories per material. For comparison, \cite{choi2024graph} used $60$ trajectories per friction angle, where the material property was embedded as a vertex feature in each particle’s input vector. These results underscore the significant computational savings of fine-tuning and conditioning based learning relative to full model retraining.

\subsection{Validation via inverse problem}

\begin{figure*}
    \includegraphics[width=\linewidth]{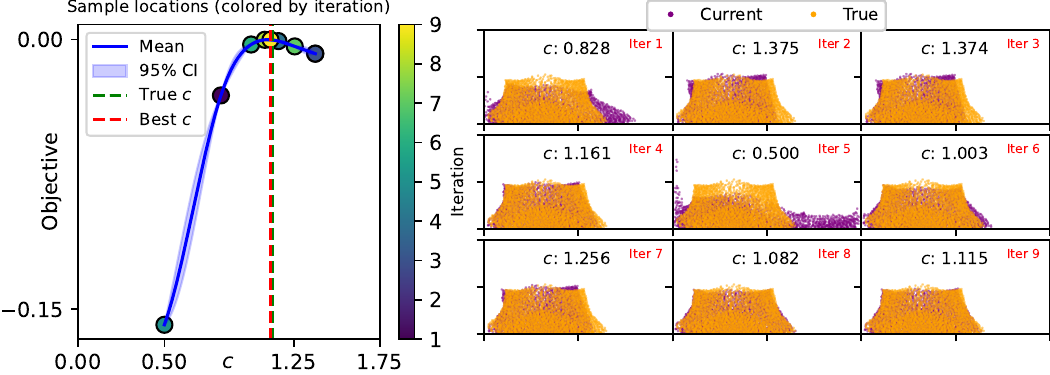}
    \caption{Bayesian optimization of the parameter $c$. The optimizer explored $c\in[0.5,\,1.375]$ directly with adaptive sampling. The loss function is MPED at terminal time step.}
    \label{fig:bayes_opt_phi}
\end{figure*}

The ability of the FiLM-conditioned GNS to generalize across material parameters also opens the door to solving a broader class of problems that would otherwise be infeasible with a pretrained, single-material model. To demonstrate this capability, we employ the FiLM-based framework as a differentiable, material-aware surrogate that can perform forward simulations over a continuous material space, thereby enabling efficient parameter inference through optimization or probabilistic search. Bayesian optimization was applied to infer the cohesion-related parameter $c$ using a ground-truth trajectory corresponding to $1.125~\mathrm{kPa}$. The optimizer immediately began adaptive sampling over the prior range $[0.5,\,1.375]$. Each forward simulation took roughly $7.3\,\mathrm{s}$ on NVIDIA’s Grace Hopper GPU. The loss is the maximum pairwise Euclidean distance (MPED) at terminal time step, defined as 
\( \mathrm{MPED}(X_T, Y_T) = \max_{i,j} \| \mathbf{x}_{i,T} - \mathbf{y}_{j,T} \|_2 \), 
where \( X_T = \{\mathbf{x}_{i,T}\} \) and \( Y_T = \{\mathbf{y}_{j,T}\} \) denote 
the particle positions in the current and target terminal states, respectively. Convergence was achieved by iteration~9 (see Fig.~\ref{fig:bayes_opt_phi}), where the loss fell to $5.81\times10^{-5}$ at $c \approx 1.1171$ kPa, accurately matching the target material.

\section{Conclusion}
\label{sec:conclusion}

This work addresses a critical limitation of graph network-based simulators (GNS): their failure to generalize to unseen material properties, which has hindered their adoption in computational mechanics. Our core contribution is the finding that this domain sensitivity is not uniformly distributed throughout the network. Instead, it is highly concentrated in the initial message-passing (MP) layers, which are architecturally suited to learn the local, material-dependent physics governing particle interactions. Building on this insight, we proposed a parameter-efficient conditioning framework that uses Feature-wise Linear Modulation (FiLM) to specifically target these early layers. This approach allows a single GNS, pretrained on data from a single material, to be efficiently adapted to a continuous spectrum of new material properties using only a small set of auxiliary data.

We further enhanced data efficiency by introducing a physics-guided sampling strategy, which identifies that the information-rich, early mobilization phase of a trajectory is sufficient for successful adaptation, reducing the high computational cost of data generation. Our experimental results, validated on a high-fidelity dataset of granular flows, show that this FiLM-conditioned GNS successfully generalizes across unseen, interpolated, and extrapolated material properties, including both internal friction angles and cohesion. The single adaptive model achieves an accuracy comparable to specialized models that are individually fine-tuned for each material. A PCA analysis of the FiLM parameters confirmed that our conditioning network learns a smooth, continuous manifold of material behaviors, enabling robust interpolation. Critically, the adapted model also demonstrates greater physical plausibility by maintaining better adherence to energy conservation.

By effectively decoupling the learning of general physical dynamics from material-specific responses, our framework makes learned simulators more practical, reliable, and efficient. We demonstrated this utility by employing the conditioned GNS as a fast, differentiable surrogate model to successfully solve an inverse problem: accurately identifying material properties from trajectory data. Future work will focus on extending this framework to jointly generalize across multi-dimensional parameter spaces (e.g., varying friction and cohesion simultaneously) and applying it to larger-scale problems. We also see significant potential in creating a tighter integration between perception models and this material-aware simulation engine, learning-accelerated data assimilation \cite{iqbal2025scalable}, and rapid inference of physical properties from observations for real-time prediction and control.

\section{Data availability}

All data and models developed in this work will be released upon publication. 

\section{Acknowledgments}
This work was funded in part by NSF 2339678, 2321040 and 2438193.  Any opinions, findings, conclusions, or recommendations expressed in this material are those of the author(s) and do not necessarily reflect the views of the funding organizations.

\clearpage

\appendix
\section{Baseline Training Stability Analysis}
\label{app_baseline}
\begin{figure}[h!]
\centering
\includegraphics{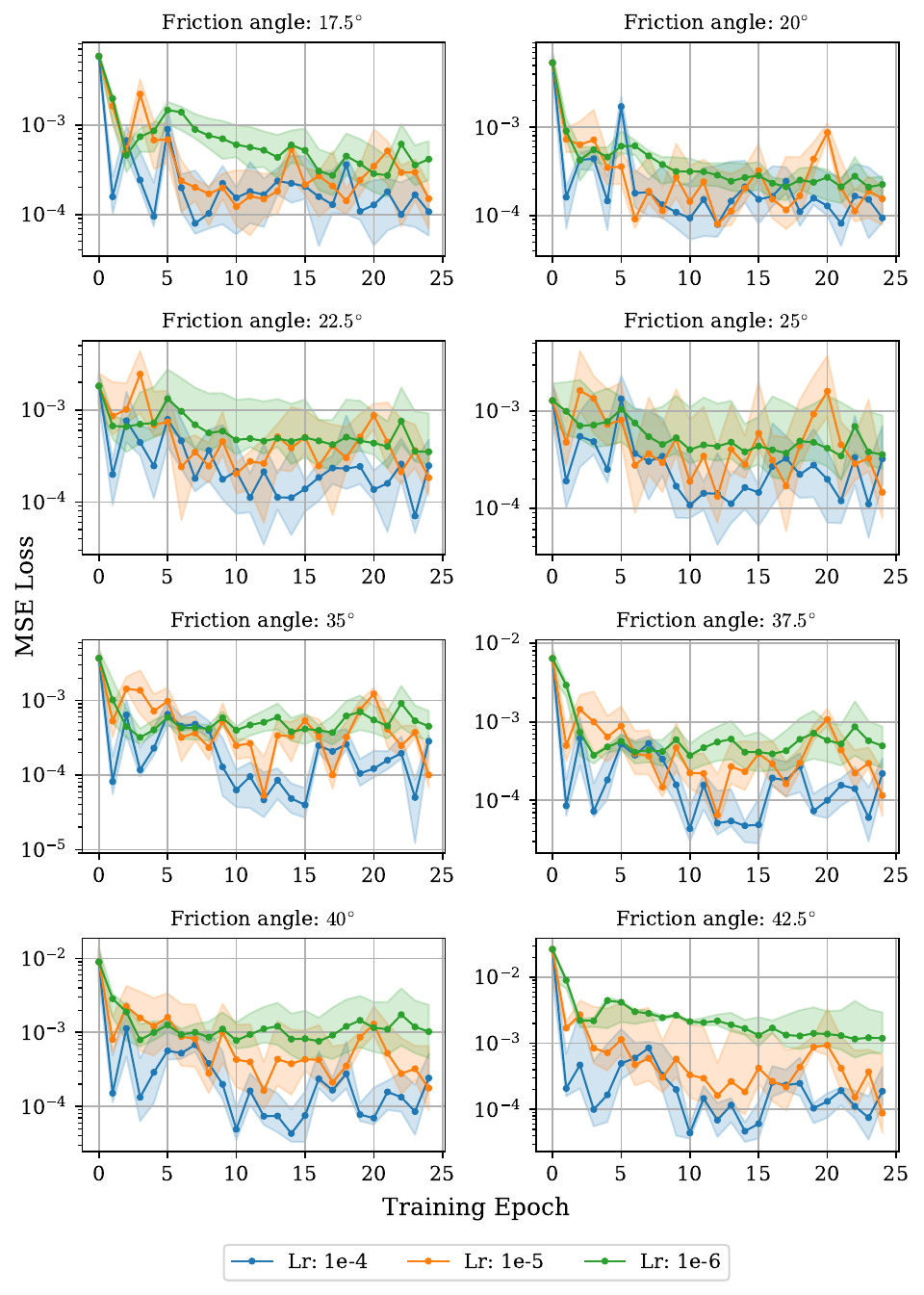}
\caption{Training stability of the baseline model for different learning rates. Each panel shows the MSE loss on test data as a function of training epoch for a specific friction angle}
\label{baseline_test_loss}
\end{figure}
To establish a reference point for comparison with the fine-tuning and FiLM-conditioned approaches discussed in the main text, we establish a baseline model. As described in Section~\ref{baseline}, this baseline model is first initialized by training from scratch using trajectories corresponding to a friction angle of $30^\circ$, after which the training dataset is expanded to include trajectories from all friction angles considered in the study.
Figure~\ref{baseline_test_loss} shows the evolution of the mean squared error (MSE) loss during training for several learning rates ($10^{-4}$, $10^{-5}$, and $10^{-6}$). Each subplot corresponds to a different friction angle used for evaluation, including training angles ($20^\circ$, $25^\circ$, $35^\circ$, and $40^\circ$), interpolated angles ($22.5^\circ$ and $37.5^\circ$), and extrapolated angles ($17.5^\circ$ and $42.5^\circ$). The curves represent the average loss across rollout trajectories, while the shaded regions indicate variability across runs.

The results demonstrate that the baseline training procedure exhibits noticeable instability across learning rates. Larger learning rates ($10^{-4}$) produce oscillatory behavior in the loss curves, while smaller learning rates reduce these oscillations at the cost of slower convergence and reduced accuracy. In contrast, the conditioning strategies proposed in the main paper exhibit substantially more stable learning dynamics.

\section{Ablation Study: Effect of Message Passing Layers}
\label{app_ablation}
\begin{figure}[ht!]
\centering
\includegraphics{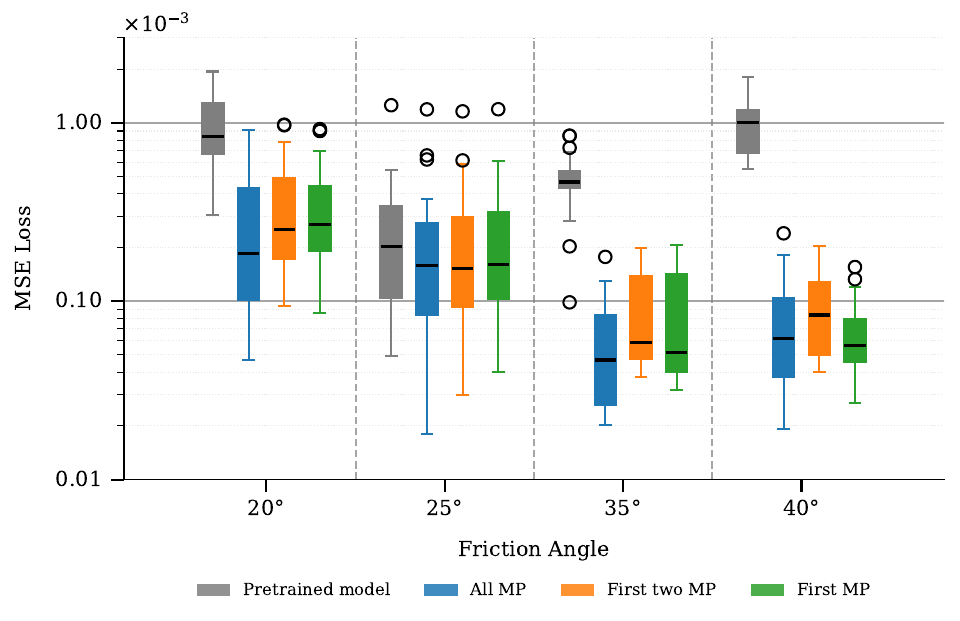}
\caption{Ablation study comparing fine-tuning strategies with different numbers of unfrozen message passing layers. Distributions show test MSE across friction angles. Partial fine-tuning achieves performance comparable to full fine-tuning with fewer trainable parameters.}
\label{ablation_study_mp}
\end{figure}
We conduct an ablation study to evaluate the impact of selectively updating message passing (MP) layers during fine-tuning. Specifically, we compare three strategies: (i) updating all MP layers, (ii) updating the first two MP layers, and (iii) updating only the first MP layer. These are evaluated against the pretrained model without fine-tuning.

Figure~\ref{ablation_study_mp} presents the distribution of test MSE across multiple friction angles. Fine-tuning consistently improves performance over the pretrained model for all configurations. Partial updates, particularly limiting adaptation to the first MP layer, yield comparable performance while reducing the number of trainable parameters and maintaining stable variance across runs. These results suggest that early MP layers capture the most transferable features for adapting to new material conditions, and that effective fine-tuning can be achieved without updating the full network.

\bibliography{refs}
\bibliographystyle{elsarticle-num}

\end{document}